\documentclass[pdflatex,sn-apa]{sn-jnl}

\usepackage{graphicx}%
\usepackage{multirow}%
\usepackage{amsmath,amssymb,amsfonts}%
\usepackage{amsthm}%
\usepackage[title]{appendix}%
\usepackage{xcolor}%
\usepackage{textcomp}%
\usepackage{manyfoot}%
\usepackage{booktabs}%
\usepackage{algorithm}%
\usepackage{algorithmicx}%
\usepackage{algpseudocode}%
\usepackage{listings}%

\usepackage{booktabs}
\usepackage{array}
\newcolumntype{L}[1]{>{\raggedright\arraybackslash}p{#1}}

\theoremstyle{thmstyleone}%
\theoremstyle{thmstyletwo}%

\theoremstyle{thmstylethree}%

\begin{document}

\title[New Money: A Systematic Review of
Synthetic Data Generation for Finance]{New Money: A Systematic Review of
Synthetic Data Generation for Finance}

\author[1]{\fnm{James} \sur{Meldrum}}\email{jmel2994@uni.sydney.edu.au}

\author*[2]{\fnm{Basem} \sur{Suleiman}}\email{b.suleiman@unsw.edu.au}

\author[2]{\fnm{Fethi} \sur{Rabhi}}\email{f.rabhi@unsw.edu.au}

\author[3]{\fnm{Muhammad Johan} \sur{Alibasa}}\email{johan.alibasa@monash.edu}

\affil[1]{\orgdiv{School of Computer Science}, \orgname{University of Sydney}, \orgaddress{\city{Sydney}, \state{NSW}, \country{Australia}}}

\affil[2]{\orgdiv{Computer Science and Engineering}, \orgname{University of New South Wales}, \orgaddress{\city{Sydney}, \state{NSW}, \country{Australia}}}

\affil[3]{\orgdiv{Faculty of Information Technology}, \orgname{Monash University Indonesia}, \orgaddress{\city{Tangerang}, \state{Banten}, \country{Indonesia}}}


\abstract{Synthetic data generation has emerged as a promising approach to address the challenges of using sensitive financial data in machine learning applications. By leveraging generative models, such as Generative Adversarial Networks (GANs) and Variational Autoencoders (VAEs), it is possible to create artificial datasets that preserve the statistical properties of real financial records while mitigating privacy risks and regulatory constraints. Despite the rapid growth of this field, a comprehensive synthesis of the current research landscape has been lacking. This systematic review consolidates and analyses 72 studies published since 2018 that focus on synthetic financial data generation. We categorise the types of financial information synthesised, the generative methods employed, and the evaluation strategies used to assess data utility and privacy. The findings indicate that GAN-based approaches dominate the literature, particularly for generating time-series market data and tabular credit data. While several innovative techniques demonstrate potential for improved realism and privacy preservation, there remains a notable lack of rigorous evaluation of privacy safeguards across studies. By providing an integrated overview of generative techniques, applications, and evaluation methods, this review highlights critical research gaps and offers guidance for future work aimed at developing robust, privacy-preserving synthetic data solutions for the financial domain.}




\maketitle

\section{Introduction}\label{sec:introduction}

Financial technology (Fintech) and the application of machine learning (ML) in the financial sector have expanded substantially over the past two decades \citep{Yeo2023AAI}. Financial institutions increasingly rely on data-driven models for credit evaluation, fraud detection, algorithmic trading, and customer management. In 2020, 83\% of financial organisations reported using machine learning within their operations \citep{horacio2019TheTech}, and spending on AI in the sector exceeded USD \$11 billion, with projections rising to USD \$31 billion by 2025 \citep{Bouzarouata2023BankingExperimentation}.

The effective use of financial data enables more informed decisions and improved services \citep{Soon2021HowServices}. However, much of this data is highly sensitive, including personal identifiers, transaction records, and credit histories. Regulatory frameworks such as the GDPR in Europe and regulations enforced by ASIC in Australia impose strict controls on how this information can be used and shared \citep{Strelcenia2023SyntheticApplicability}.

Synthetic data generation, which leverages generative models to produce artificial datasets, offers a promising approach to address these challenges \citep{2021WhatData}. Synthetic datasets can replicate the statistical properties of real data while reducing privacy risks and enabling broader sharing and experimentation. In practice, generative models can create nearly unlimited quantities of realistic data that are unlinked to any specific individuals. Despite these advantages, research into synthetic data has historically focused on text and image generation, particularly in healthcare. Comprehensive analyses of synthetic data generation techniques applied to financial datasets remain limited.

This observation motivates the present study. Specifically, we aim to address the following research questions:

\begin{enumerate}
    \item What types of financial data have been synthesised in the current literature?
    \item Which generative models have been employed for synthetic financial data generation?
    \item What evaluation methods have been used to assess the quality and privacy of synthetic datasets?
\end{enumerate}

The contributions of this review are threefold. First, we provide an exhaustive synthesis of research on synthetic financial data generation published since 2018. Second, we critically analyse the generative techniques applied to a range of data types and financial tasks. Third, we review evaluation practices to inform standardisation efforts and highlight areas for further research.

\section{Related Work}

\subsection{Background and Key Concepts}

Synthetic data generation refers to the use of generative models to create artificial datasets that replicate important statistical properties of original data while reducing privacy risks \citep{2021WhatData}. In the financial sector, such datasets can support model development, address class imbalance, and enable compliant data sharing.

Two main classes of generative models are commonly applied in this domain. Generative Adversarial Networks (GANs) consist of a generator and a discriminator trained adversarially to produce realistic samples \citep{Goodfellow2014GAN}. Variational Autoencoders (VAEs) encode data into latent probabilistic representations and reconstruct synthetic samples from this space \citep{Kingma2013Auto-EncodingBayes}. Other techniques, including style transfer and privacy-preserving frameworks such as Private Aggregation of Teacher Ensembles (PATE), have been investigated in specific contexts but remain less widely adopted in finance.

Synthetic datasets in finance are typically tabular or time-series. Evaluation criteria commonly include statistical similarity to real data distributions, machine learning efficacy (e.g., predictive performance), and privacy preservation (e.g., preventing re-identification). This subsection provides essential context for understanding the generative methods and evaluation strategies assessed in the remainder of this review.

\subsection{Synthetic Data Generation for Finance}

A growing body of research has examined the use of synthetic data generation to address privacy, regulatory, and technical challenges in financial machine learning. Several studies have highlighted that financial data are among the most sensitive forms of information, subject to strict legal requirements such as the GDPR and requiring robust privacy safeguards during analysis and model development \citep{Assefa2021GeneratingPitfalls, Strelcenia2023SyntheticApplicability}. 

Early work in this area often focused on describing motivations and outlining potential benefits, including improved data sharing, mitigation of class imbalance, and enhanced machine learning performance \citep{Assefa2021GeneratingPitfalls}. More recent studies have explored specific generative techniques, with Generative Adversarial Networks (GANs) and Variational Autoencoders (VAEs) emerging as the most widely adopted methods for synthesising tabular and time-series financial data \citep{Strelcenia2023SyntheticApplicability, Eckerli2021GENERATIVEPREPRINT, Singh2022AnApplications}. 

Although GAN-based approaches have demonstrated promising results, including realistic synthetic samples for training predictive models and simulating trading activity, their applications often lack rigorous evaluation of privacy preservation and data utility \citep{Eckerli2021GENERATIVEPREPRINT, Jordon2022SyntheticHow}. Additionally, VAEs, while theoretically well-suited for generating structured data, appear to be infrequently used in financial contexts, with only a few studies applying them to stock option data or credit risk modelling \citep{Singh2022AnApplications}.

Beyond the technical aspects, several publications have discussed practical and regulatory considerations when adopting synthetic data in financial organisations. For example, authors have emphasised the importance of clear policies on data retention and sharing, as well as mechanisms to ensure stakeholder trust and regulatory compliance \citep{James2021SyntheticUtility}. However, these contributions typically stop short of offering detailed frameworks or comparative evaluations of generative models in financial settings.

\subsection{Reviews of Synthetic Data Generation}

More general surveys of synthetic data generation have been published across domains, especially in healthcare and image analysis. Some reviews have outlined a broad taxonomy of methods, including GANs, VAEs, and hybrid approaches, but tend to focus primarily on computer vision tasks \citep{Figueira2022SurveyGANs}. Others have discussed the potential of synthetic data to mitigate data scarcity and enhance machine learning workflows, while acknowledging that domain-specific challenges remain underexplored \citep{Lu2023MachineReview, Abufadda2021ALearning}.

For instance, \citet{Hernandez2022SyntheticReview} conducted a systematic review of synthetic data generation for tabular health records and found that GANs generally outperform other techniques in terms of statistical similarity and model training efficacy. However, their analysis also revealed a lack of standardised metrics for assessing privacy and data resemblance, a limitation echoed in several other studies \citep{Reiter2023SyntheticForward}. 

In financial applications, existing reviews have primarily provided high-level overviews without a comprehensive, structured comparison of methods and evaluation strategies \citep{Kharkiv2023Editura2023, Jordon2022SyntheticHow}. This gap highlights the need for a focused synthesis of synthetic data generation techniques and practices specific to finance, which is the aim of this work.

\section{Methodology}\label{sec:methodology}

\begin{figure}
    \centering
    \includegraphics[width=\textwidth]{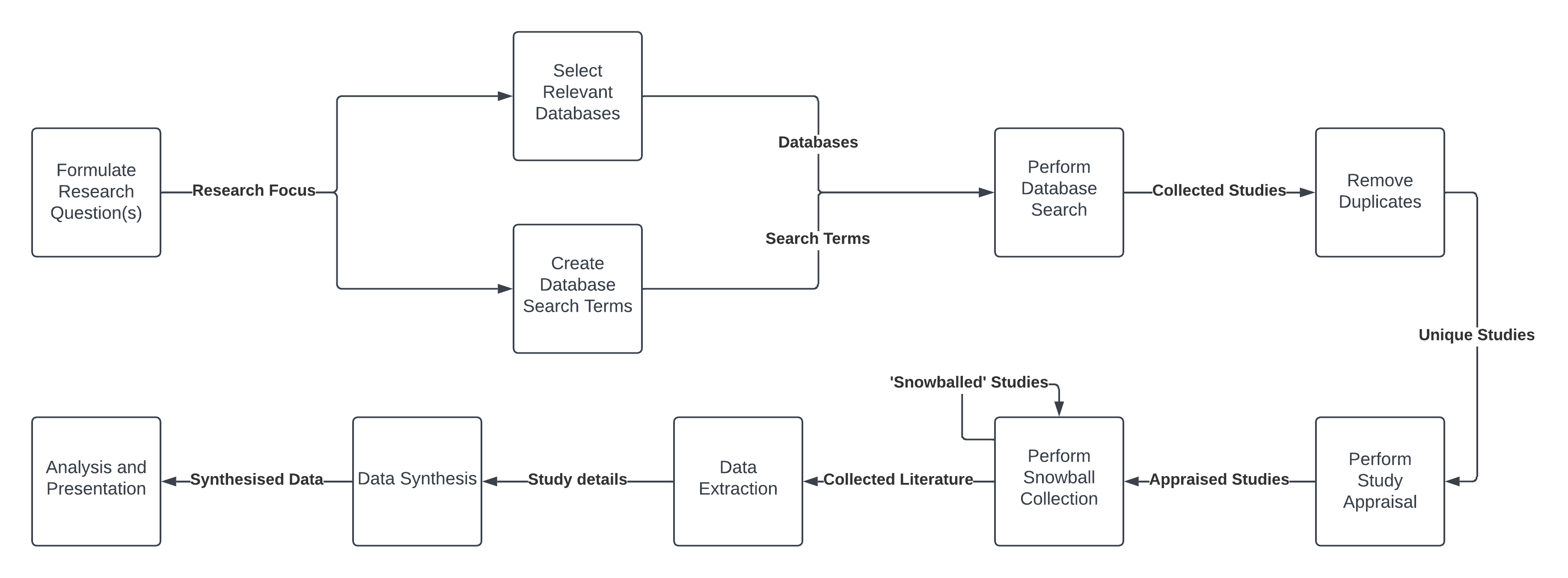}
    \caption{Overview of the systematic review process.}
    \label{fig:slr-process}
\end{figure}

This review critically analyses the current state of research on synthetic data generation for financial applications. Following the Preferred Reporting Items for Systematic Reviews and Meta-Analyses (PRISMA) guidelines \citep{PRISMA2009}, the review was conducted in four stages, illustrated in Figure~\ref{fig:slr-process}. First, a two-stage search strategy was developed across five research databases. Second, the search was executed to identify and screen studies relevant to the research questions. Third, data were extracted from the included studies. Finally, the extracted information was synthesised and analysed.

\subsection{Search Strategy}

To identify all potentially relevant studies, the search was conducted in two phases: a database search and a snowball search.

\subsubsection{Database Search}

Five databases were queried:

\begin{enumerate}
    \item ACM Digital Library
    \item IEEE Xplore
    \item Scopus
    \item SpringerLink
    \item Web of Science
\end{enumerate}

Search strings combined keywords related to synthetic data generation and finance. A study was considered for inclusion if at least one keyword from each category appeared in the title, abstract, or keyword list:

\begin{enumerate}
    \item Synthetic Data Generation:
    \begin{itemize}
        \item ``data generat*''
        \item ``synthetic data*''
        \item ``generated data*''
        \item ``artificial data*''
    \end{itemize}
    \item Finance:
    \begin{itemize}
        \item ``financ*''
        \item ``econom*''
        \item ``bank*''
        \item ``stock*''
    \end{itemize}
\end{enumerate}

Where possible, filters were applied to limit results by publication date and language.

\subsubsection{Snowball Search}

After the database search, forward and backward snowballing was performed \citep{Wohlin2014GuidelinesEngineering}. References cited by included studies (backward snowballing) and studies citing them (forward snowballing) were reviewed iteratively until no new relevant publications were identified. All studies identified through snowballing were screened with the same criteria as the database search.

\subsection{Screening Strategy}

Screening was conducted in two stages to assess relevance and quality.

Prior to screening, two exclusion criteria were enforced using database filters:

\begin{itemize}
    \item Studies published before 2018 were excluded to ensure coverage of recent advancements.
    \item Studies not published in English were excluded due to language limitations.
\end{itemize}

\subsubsection{Title and Abstract Screening}

In the first phase, titles and abstracts were reviewed to exclude clearly irrelevant studies. The criteria applied were:

\begin{itemize}
    \item Studies explicitly focused on fields unrelated to finance or computer science were excluded.
    \item Studies that did not mention synthetic data generation or related terms were excluded.
    \item Studies that did not mention a research focus relevant to finance or computer science were excluded.
    \item Studies that explicitly described the generation of exclusively non-financial data were excluded.
\end{itemize}

\subsubsection{Full-Text Screening}

Remaining studies were assessed in full text against the following criteria:

\begin{itemize}
    \item Studies must describe the generation of financial (or closely related) data.
    \item Studies must protect sensitive or personally identifiable information.
    \item Studies must state the data generation method used.
    \item Studies generating data not based on existing datasets were excluded.
    \item Studies focused exclusively on minority oversampling, forecasting, or unrelated machine learning tasks were excluded.
\end{itemize}

\subsection{Data Extraction}

For each included study, data were extracted to address the research questions. Table~\ref{tab:methodology-data-points} summarises the data points collected.

\begin{table}
\centering
\renewcommand{\arraystretch}{1.2}
\begin{tabular}{L{1.5cm} p{10cm}}
\toprule
\textbf{Research Question} & \textbf{Data Points Extracted} \\
\midrule
RQ1 & Data types synthesised (e.g., time series, tabular); financial applications targeted (e.g., stock exchange, transactions). \\
RQ2 & Type of generative techniques used; specific implementations. \\
RQ3 & Evaluation focus (statistical similarity, machine learning efficacy, privacy preservation); metrics and methods employed (e.g., visual inspection, F1 score, comparisons to baselines). \\
\bottomrule
\end{tabular}
\caption{Summary of data points extracted from included studies.}
\label{tab:methodology-data-points}
\end{table}

\subsection{Data Synthesis}

Data synthesis involved categorising and analysing the extracted data by attributes including generative techniques, applications, evaluation methods, and publication year. Results were organised into tables and visualisations to support interpretation and discussion. Data processing and analysis were conducted using Microsoft Excel and Python, with libraries including Pandas, Matplotlib, and Plotly.

\section{Results}

\subsection{Study Selection and Collection}

\begin{figure}
    \centering
    \fbox{\includegraphics[width=\textwidth]{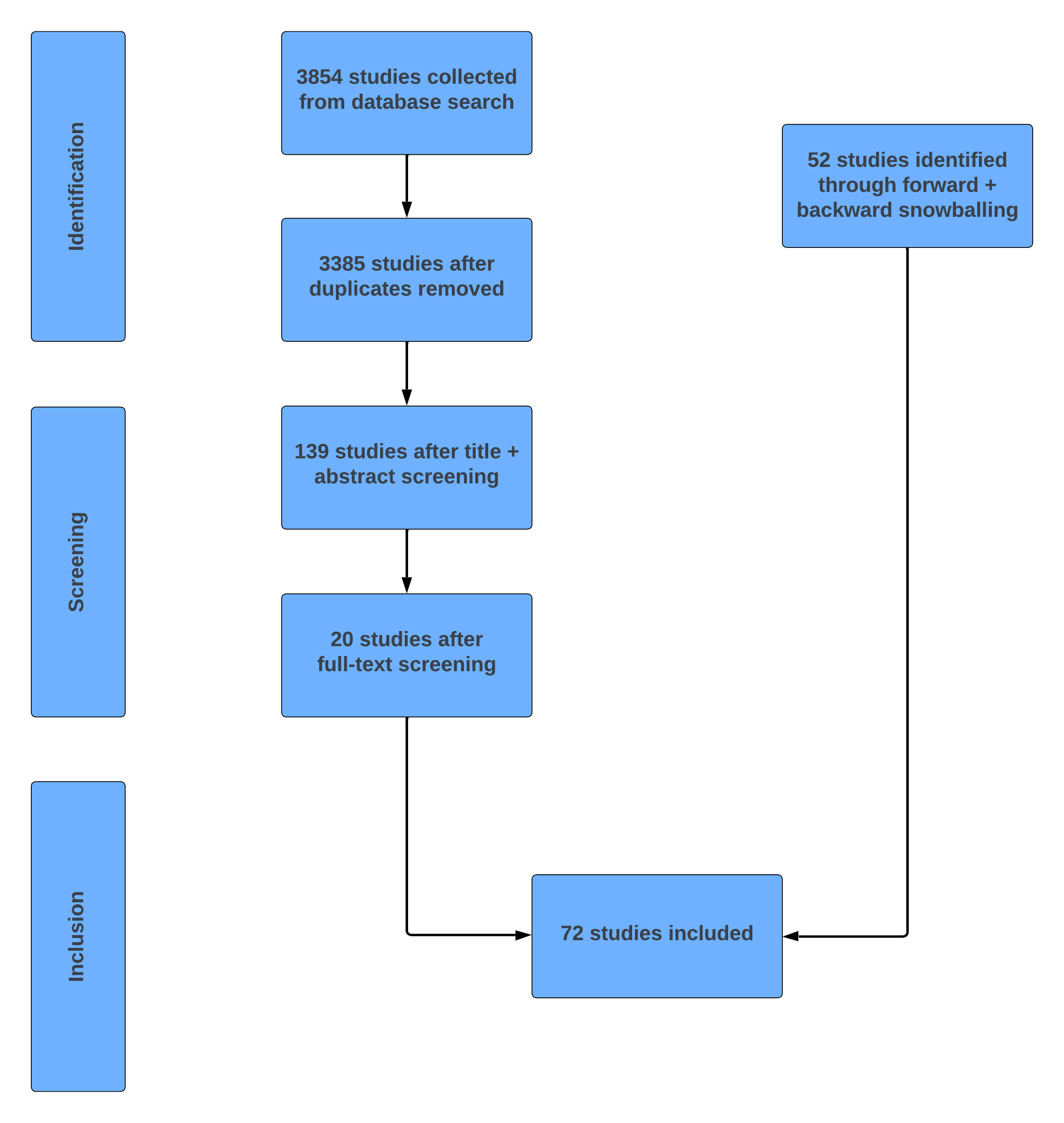}}
    \caption{PRISMA flow diagram.}
    \label{fig:prisma-flow}
\end{figure}

In total, we collected 72 studies focused on the generation of synthetic financial data across a diverse range of applications within the industry. Figure~\ref{fig:prisma-flow} illustrates the process of study identification, screening, and inclusion, as described in Section~\ref{sec:methodology}. From the initial retrieval of studies, the majority were excluded as irrelevant to the review’s focus, with 3,246 records removed prior to full-text screening.

Notably, most of the included studies were identified during the snowball search phase rather than through the initial database queries. This likely reflects a gap between the terminology used in search strategies, where finance-specific keywords were essential, and the way many authors report their research. In many cases, financial applications of synthetic data were only mentioned within the methodology sections rather than in titles, abstracts, or keywords.

\begin{figure}
    \centering
    \includegraphics[width=\textwidth]{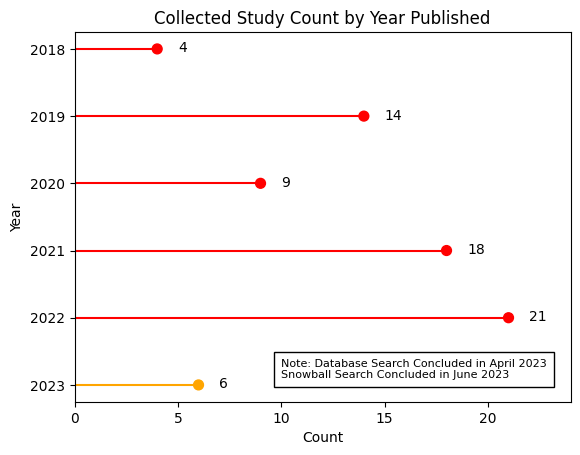}
    \caption{Number of studies collected by published year.}
    \label{fig:lollypop}
\end{figure}

The number of relevant publications has grown steadily over recent years, as shown in Figure~\ref{fig:lollypop}. One exception is 2020, when only nine studies were published, potentially due to the disruption caused by the COVID-19 pandemic. The relatively lower count in 2023 is attributable to the data collection occurring during the first half of that year. Overall, these results indicate a consistent and increasing research interest in the use of synthetic data generation for financial applications.

\begin{table}
\centering
\renewcommand{\arraystretch}{1.2}
\begin{tabular}{L{0.4cm} L{2.3cm} L{6.7cm} L{2cm}}
\toprule
\textbf{ID} & \textbf{Citation} & \textbf{Title} & \textbf{Task} \\
\midrule
14 & \cite{Kegel2018Feature-basedSeries} & Feature-Based Comparison and Generation of Time Series & Not Specified \\
30 & \cite{Park2018DataNetworks} & Data Synthesis Based on Generative Adversarial Networks & Retail Prices \\
34 & \cite{Xiao2018LearningProcesses} & Learning Conditional Generative Models for Temporal Point Processes & Stock Market \\
39 & \cite{Simonetto2018GeneratingTransactions} & Generating Spiking Time Series with Generative Adversarial Networks: An Application on Banking Transactions & Transaction \\
\bottomrule
\end{tabular}
\caption{Studies collected from 2018.}
\label{tab:2018-studies}
\end{table}

\begin{table}
\centering
\renewcommand{\arraystretch}{1.2}
\begin{tabular}{L{0.4cm} L{2.3cm} L{6.7cm} L{2cm}}
\toprule
\textbf{ID} & \textbf{Citation} & \textbf{Title} & \textbf{Task} \\
\midrule
12 & \cite{Raimbault2019Second-orderData} & Second-Order Control of Complex Systems with Correlated Synthetic Data & Currency Exchange \\
13 & \cite{Zhang2019StockNetwork} & Stock Market Prediction Based on Generative Adversarial Network & Stock Market \\
17 & \cite{Jordon2019PATE-GAN:GUARANTEES} & Pate-GAN: Generating Synthetic Data with Differential Privacy Guarantees & Credit \\
18 & \cite{Koshiyama2019GenerativeCombination} & Generative Adversarial Networks for Financial Trading Strategies Fine-Tuning and Combination & Currency Exchange, Stock Market \\
20 & \cite{Wiese2019QuantSeries} & Quant GANs: Deep Generation of Financial Time Series & Stock Market \\
21 & \cite{Yoon2019Time-seriesNetworks} & Time-Series Generative Adversarial Networks & Stock Market \\
24 & \cite{DaSilva2019StyleData} & Style Transfer with Time Series: Generating Synthetic Financial Data & Currency Exchange \\
27 & \cite{Fu2019TimeNet} & Time Series Simulation by Conditional Generative Adversarial Net & Stock Market \\
28 & \cite{Abay2019PrivacyLearning} & Privacy Preserving Synthetic Data Release Using Deep Learning & Credit \\
32 & \cite{Xu2019ModelingGAN} & Modeling Tabular Data Using Conditional GAN & Credit \\
36 & \cite{De2019EnrichingNetworks} & Enriching Financial Datasets with Generative Adversarial Networks & Stock Market \\
38 & \cite{Brenninkmeijer2019OnGANs} & On the Generation and Evaluation of Tabular Data Using GANs & Credit, Transaction \\
66 & \cite{Miok2019GeneratingDropout} & Generating Data Using Monte Carlo Dropout & Credit \\
67 & \cite{Takahashi2019ModelingNetworks} & Modeling Financial Time Series with Generative Adversarial Networks & Stock Market \\
\bottomrule
\end{tabular}
\caption{Studies collected from 2019.}
\label{tab:2019-studies}
\end{table}

\begin{table}
\centering
\renewcommand{\arraystretch}{1.2}
\begin{tabular}{L{0.4cm} L{2.3cm} L{6.7cm} L{2cm}}
\toprule
\textbf{ID} & \textbf{Citation} & \textbf{Title} & \textbf{Task} \\
\midrule
11 & \cite{Zhang2020ASamples} & A Generative Adversarial Network Based Method for Generating Negative Financial Samples & Transaction \\
19 & \cite{Li2020GeneratingStreams} & Generating Realistic Stock Market Order Streams & Market Orders \\
22 & \cite{Efimov2020UsingDatasets} & Using Generative Adversarial Networks to Synthesize Artificial Financial Datasets & Not Specified \\
25 & \cite{vanBree2020UnlockingAutoencoders} & Unlocking the Potential of Synthetic Tabular Data Generation with Variational Autoencoders & Credit \\
31 & \cite{Padhi2020TabularSeries} & Tabular Transformers for Modeling Multivariate Time Series & Transaction \\
33 & \cite{Kondratyev2020TheGenerator} & The Market Generator & Currency Exchange \\
37 & \cite{Karlsson2020SynthesisNetworks} & Synthesis of Tabular Financial Data Using Generative Adversarial Networks & Marketing \\
42 & \cite{Vega-Marquez2020CreationNetworks} & Creation of Synthetic Data with Conditional Generative Adversarial Networks & Credit \\
65 & \cite{Goos2020Privacy-PreservingData} & Privacy-Preserving Anomaly Detection Using Synthetic Data & Transaction \\
\bottomrule
\end{tabular}
\caption{Studies collected from 2020.}
\label{tab:2020-studies}
\end{table}

\begin{table}
\centering
\renewcommand{\arraystretch}{1.2}
\begin{tabular}{L{0.4cm} L{2.3cm} L{6.7cm} L{2cm}}
\toprule
\textbf{ID} & \textbf{Citation} & \textbf{Title} & \textbf{Task} \\
\midrule
2 & \cite{Coletta2021TowardsApproach} & Towards Realistic Market Simulations: A Generative Adversarial Networks Approach & Market Orders \\
9 & \cite{Ni2020ConditionalGeneration} & Sig-Wasserstein GANs for Time Series Generation & Stock Market \\
10 & \cite{Park2021SynthesizingNetworks} & Synthesizing Individual Consumers' Credit Historical Data Using Generative Adversarial Networks & Loan, Credit \\
23 & \cite{Desai2021TimeVAE:Generation} & TimeVAE: A Variational Auto-Encoder for Multivariate Time Series Generation & Stock Market \\
26 & \cite{Dogariu2021TowardsLearning} & Towards Realistic Financial Time Series Generation via Generative Adversarial Learning & Stock Market \\
29 & \cite{Ljung2021SyntheticNetworks} & Synthetic Data Generation for the Financial Industry Using Generative Adversarial Networks & Marketing \\
35 & \cite{Zhao2021CTAB-GAN:Synthesizing} & CTAB-GAN: Effective Table Data Synthesizing & Loan, Credit \\
43 & \cite{Kim2021OCT-GAN:GANs} & OCT-GAN: Neural ODE-Based Conditional Tabular GANs & Credit \\
44 & \cite{Long2021G-PATE:Discriminators} & G-PATE: Scalable Differentially Private Data Generator via Private Aggregation of Teacher Discriminators & Credit \\
45 & \cite{Li2021ImprovingSynthesis} & Improving GAN with Inverse Cumulative Distribution Function for Tabular Data Synthesis & Credit \\
46 & \cite{VanBreugel2021DECAF:Networks} & DECAF: Generating Fair Synthetic Data Using Causally-Aware Generative Networks & Credit \\
52 & \cite{Remlinger2021ConditionalGeneration} & Conditional Loss and Deep Euler Scheme for Time Series Generation & Stock Market, Other \\
56 & \cite{Li2021AG-XGBoost} & A Credit Risk Model with Small Sample Data Based on G-XGBoost & Credit \\
59 & \cite{Pei2021TowardsData} & Towards Generating Real-World Time Series Data & Stock Market \\
60 & \cite{Yin2021Multi-AttentionPrediction} & Multi-Attention Generative Adversarial Network for Multivariate Time Series Prediction & Stock Market \\
68 & \cite{Alaa2021GENERATIVEFLOWS} & Generative Time-Series Modeling with Fourier Flows & Stock Market \\
73 & \cite{Platzer2021Holdout-BasedData} & Holdout-Based Empirical Assessment of Mixed-Type Synthetic Data & Credit, Other \\
74 & \cite{Ge2021Kamino:Synthesis} & Kamino: Constraint-Aware Differentially Private Data Synthesis & Tax \\
\bottomrule
\end{tabular}
\caption{Studies collected from 2021.}
\label{tab:2021-studies}
\end{table}

\begin{table}
\centering
\renewcommand{\arraystretch}{1.2}
\begin{tabular}{L{0.4cm} L{3cm} L{8cm} L{3cm}}
\toprule
\textbf{ID} & \textbf{Citation} & \textbf{Title} & \textbf{Task} \\
\midrule
0 & \cite{Liu2022SyntheticTrading} & Synthetic Data Augmentation for Deep Reinforcement Learning in Financial Trading & Stock Market \\
1 & \cite{El-Laham2022StyleTime:Generation} & StyleTime: Style Transfer for Synthetic Time Series Generation & Stock Market \\
5 & \cite{Dogariu2022GenerationTime-series} & Generation of Realistic Synthetic Financial Time-Series & Stock Market \\
6 & \cite{Coletta2022LearningAgent} & Learning to Simulate Realistic Limit Order Book Markets from Data as a World Agent & Market Orders \\
7 & \cite{Azamuke2022Scenario-basedTransactions} & Scenario-Based Synthetic Dataset Generation for Mobile Money Transactions & Transaction \\
8 & \cite{Rizzato2022StressGeneration} & Stress Testing Electrical Grids: Generative Adversarial Networks for Load Scenario Generation & Commodities \\
16 & \cite{Vega-Marquez2022GenerationNetworks} & Generation of Synthetic Data with Conditional Generative Adversarial Networks & Credit \\
41 & \cite{Tan2022DeepPricing:Networks} & DeepPricing: Pricing Convertible Bonds Based on Financial Time-Series Generative Adversarial Networks & Stock Market \\
47 & \cite{Lee2022InvertibleSynthesis} & Invertible Tabular GANs: Killing Two Birds with One Stone for Tabular Data Synthesis & Credit \\
48 & \cite{Duan2022HT-Fed-GAN:Synthesis} & HT-Fed-GAN: Federated Generative Model for Decentralized Tabular Data Synthesis & Credit \\
50 & \cite{Nickerson2022Banksformer:Sequences} & Banksformer: A Deep Generative Model for Synthetic Transaction Sequences & Transaction \\
53 & \cite{Flaig2022ScenarioNetworks} & Scenario Generation for Market Risk Models Using Generative Neural Networks & Economic Scenario \\
54 & \cite{Allouche2022EV-GAN:Networks} & EV-GAN: Simulation of Extreme Events with ReLU Neural Networks & Stock Market \\
55 & \cite{Hayashi2022FractionalMemory} & Fractional SDE-Net: Generation of Time Series Data with Long-Term Memory & Stock Market \\
61 & \cite{Geissler2022GenerativeGeneration} & Generative Adversarial Networks Applied to Synthetic Financial Scenarios Generation & Currency Exchange, Commodities, Credit, Stock Market \\
62 & \cite{Gatta2022NeuralSeries} & Neural Networks Generative Models for Time Series & Stock Market \\
63 & \cite{Jeon2022GT-GAN:Networks} & GT-GAN: General Purpose Time Series Synthesis with Generative Adversarial Networks & Stock Market \\
64 & \cite{Juneja2023SyntheticAnalysis} & Synthetic Time Series Data Generation Using TimeGAN with Synthetic and Real-Time Data Analysis & Stock Market \\
69 & \cite{Cramer2022ValidationModels} & Validation Methods for Energy Time Series Scenarios from Deep Generative Models & Commodities \\
71 & \cite{CarvajalPatino2022SyntheticStrategies} & Synthetic Data Generation with Deep Generative Models to Enhance Predictive Tasks in Trading Strategies & Commodities, Currency Exchange \\
72 & \cite{Boursin2022DeepHedging} & Deep Generators on Commodity Markets: Application to Deep Hedging & Commodities \\
\bottomrule
\end{tabular}
\caption{Studies collected from 2022.}
\label{tab:2022-studies}
\end{table}

\begin{table}
\centering
\renewcommand{\arraystretch}{1.2}
\begin{tabular}{L{0.4cm} L{2.3cm} L{6.7cm} L{2cm}}
\toprule
\textbf{ID} & \textbf{Citation} & \textbf{Title} & \textbf{Task} \\
\midrule
15 & \cite{Zhang2023InterpretableGeneration} & Interpretable Tabular Data Generation & Loan, Credit \\
40 & \cite{Yadav2023QualitativeApproach} & Qualitative and Quantitative Evaluation of Multivariate Time-Series Synthetic Data Generated Using MTS-TGAN: A Novel Approach & Stock Market \\
51 & \cite{Wu2023AApproaches} & A Prediction Model of Stock Market Trading Actions Using Generative Adversarial Network and Piecewise Linear Representation Approaches & Stock Market \\
57 & \cite{Tang2023AForecasting} & A Recurrent Neural Network Based Generative Adversarial Network for Long Multivariate Time Series Forecasting & Currency Exchange \\
58 & \cite{Ahmed2023SparseGeneration} & Sparse Self-Attention Guided Generative Adversarial Networks for Time-Series Generation & Stock Market \\
70 & \cite{Wu2023InterpretationData} & Interpretation for Variational Autoencoder Used to Generate Financial Synthetic Tabular Data & Other, Loan \\
\bottomrule
\end{tabular}
\caption{Studies collected from 2023.}
\label{tab:2023-studies}
\end{table}

\subsection{What financial information has been synthesised throughout the relevant literature?} 
\label{rq:1}

\subsubsection{Market Data}

We find the generation of univariate or multivariate stock market data to be the most common application of synthetic data generation within our collected studies. Most studies synthesising stock market data generated a combination of (or all of) daily opening, closing, high and low stock prices, adjusted closing prices, volume, and turnover rate for one or multiple stocks and indexes. We note that for studies generating univariate stock prices, we recorded this as daily closing prices unless stated otherwise. Many of these studies used the synthetic market data to train machine learning models such as trading agents or market price forecasting systems.

Market order information involves the synthesis of data representing stock order streams (buy/sell signals, price, volume, and similar features). \cite{Coletta2021TowardsApproach} and \cite{Coletta2022LearningAgent} used a single generative model to produce order streams for the entire market that react to the activity of experimental agents, as an alternative to simulating many trading agents independently. This approach enables the creation of realistic market scenarios for testing trading strategies. \cite{Li2020GeneratingStreams} similarly generated market order streams with historical dependencies, aiming to improve the ability to analyse sensitive stock market information.

A related time-series application is the synthesis of exchange information. A number of the collected studies synthesised correlated exchange rates between currencies of two or more countries. For example, \cite{DaSilva2019StyleData} generated realistic exchange rates between AUD and USD. \cite{Boursin2022DeepHedging} similarly produced correlated prices of coal, gas, electricity, and oil to perform hedging on futures contracts using deep learning. \cite{CarvajalPatino2022SyntheticStrategies} synthesised both currency and commodity data in the form of correlated exchange rates between the price of gold, USD, and EUR.

\begin{table}
\centering
\begin{tabular}{@{}L{2cm}L{2cm}L{1cm}L{6cm}@{}}
\toprule
\textbf{Category} & \textbf{Task} & \textbf{Count} & \textbf{Studies ID} \\ \midrule
\multirow{10}{*}{Stock Market} 
& Open & 13 & {[}0{]} {[}13{]} {[}21{]} {[}23{]} {[}26{]} {[}40{]} {[}51{]} {[}52{]} {[}58{]} {[}59{]} {[}62{]} {[}63{]} {[}64{]} \\
& Close & 25 & {[}0{]} {[}1{]} {[}5{]} {[}9{]} {[}13{]} {[}18{]} {[}20{]} {[}21{]} {[}23{]} {[}26{]} {[}27{]} {[}36{]} {[}40{]} {[}51{]} {[}52{]} {[}54{]} {[}55{]} {[}58{]} {[}59{]} {[}60{]} {[}61{]} {[}62{]} {[}63{]} {[}64{]} {[}67{]} \\
& High & 13 & {[}0{]} {[}13{]} {[}21{]} {[}23{]} {[}26{]} {[}40{]} {[}51{]} {[}52{]} {[}58{]} {[}59{]} {[}62{]} {[}63{]} {[}64{]} \\
& Low & 13 & {[}0{]} {[}13{]} {[}21{]} {[}23{]} {[}26{]} {[}40{]} {[}51{]} {[}52{]} {[}58{]} {[}59{]} {[}62{]} {[}63{]} {[}64{]} \\
& Adjusted Close & 10 & {[}0{]} {[}21{]} {[}23{]} {[}40{]} {[}51{]} {[}52{]} {[}58{]} {[}59{]} {[}63{]} {[}64{]} \\
& Volume & 12 & {[}0{]} {[}13{]} {[}21{]} {[}23{]} {[}40{]} {[}51{]} {[}52{]} {[}58{]} {[}59{]} {[}62{]} {[}63{]} {[}64{]} \\
& Turnover Rate & 1 & {[}13{]} \\
& 5-Day Average & 1 & {[}13{]} \\
& Transactions & 1 & {[}34{]} \\
& Details Not Specified & 2 & {[}41{]} {[}68{]} \\ \midrule
Market Orders & Market Orders & 3 & {[}2{]} {[}6{]} {[}19{]} \\ \midrule
Currency Exchange & Currency Exchange & 7 & {[}12{]} {[}18{]} {[}24{]} {[}33{]} {[}57{]} {[}61{]} {[}71{]} \\ \midrule
Commodities & Commodities & 5 & {[}8{]} {[}61{]} {[}69{]} {[}71{]} {[}72{]} \\ \bottomrule
\end{tabular}
\caption{Studies synthesising market data.}
\label{tab:market-papers}
\end{table}

\subsubsection{Credit and Loan Data}

A large portion of the literature also focuses on synthesising credit and loan data. As opposed to market data, which is mostly time series, credit data is primarily mixed-type tabular data. A common use case is the detection of fraudulent behaviours. We found that nine studies generated synthetic credit and loan data for this purpose. The other most frequent application was assessing customer credit risk. This is consistent with the fact that personal credit information is highly sensitive, and the ability to use synthetic versions without risking privacy breaches is valuable. A similar motivation applies to the four studies that synthesised personal loan data.

\begin{table}
\centering
\begin{tabular}{@{}L{2cm}L{2cm}L{1cm}L{6cm}@{}}
\toprule
\textbf{Category} & \textbf{Task} & \textbf{Count} & \textbf{Studies ID} \\ \midrule
\multirow{5}{*}{Credit}
& Credit Card & 1 & {[}10{]} \\
& Credit Risk & 9 & {[}10{]} {[}16{]} {[}28{]} {[}46{]} {[}47{]} {[}56{]} {[}61{]} {[}66{]} {[}73{]} \\
& Credit (Not Specified) & 2 & {[}15{]} {[}45{]} \\
& Credit Fraud & 9 & {[}17{]} {[}25{]} {[}32{]} {[}35{]} {[}38{]} {[}42{]} {[}43{]} {[}44{]} {[}48{]} \\
& Personal Loan & 4 & {[}10{]} {[}15{]} {[}35{]} {[}70{]} \\ \bottomrule
\end{tabular}
\caption{Studies synthesising credit and loan data.}
\label{tab:credit-loan-papers}
\end{table}

\subsubsection{Other Applications}

Among the remaining applications, the generation of synthetic transaction data was the most common, appearing in seven studies. Four studies generated marketing and customer churn data for banking institutions. \cite{Flaig2022ScenarioNetworks} created synthetic economic scenarios for insurance risk calculations. Interestingly, only one study generated synthetic tax data, which represents a potentially important area for future work given the sensitivity of such records.

\begin{table}
\centering
\begin{tabular}{@{}L{2cm}L{2cm}L{1cm}L{6cm}@{}}
\toprule
\textbf{Category} & \textbf{Task} & \textbf{Count} & \textbf{Studies ID} \\ \midrule
\multirow{7}{*}{Other}
& Transactions & 7 & {[}7{]} {[}11{]} {[}31{]} {[}38{]} {[}39{]} {[}50{]} {[}65{]} \\
& Not Specified & 2 & {[}14{]} {[}22{]} \\
& Marketing & 4 & {[}29{]} {[}37{]} {[}70{]} {[}73{]} \\
& Churn & 2 & {[}29{]} {[}70{]} \\
& Retail Prices & 2 & {[}30{]} {[}52{]} \\
& Economic Scenario & 1 & {[}53{]} \\
& Tax & 1 & {[}74{]} \\ \bottomrule
\end{tabular}
\caption{Studies applying synthetic data in other financial domains.}
\label{tab:other-apps-papers}
\end{table}

Overall, we find that the main applications of synthetic financial data in the literature are in stock and market data generation, credit risk, and credit fraud detection. Opportunities for future work include further generation of transaction, retail, and tax data to broaden the applicability of synthetic data across financial institutions.

\subsection{What generative models have been used for the generation of financial data?}
\label{rq:2}

To answer this research question, we isolated the studies that contained experiments assessing generative techniques for synthetic data generation. We summarised the methods discussed across these studies in five groups, based on the taxonomy illustrated in Figure~\ref{fig:models-sundial}: Conditional GANs, Vanilla and Wasserstein GANs, Other GANs, Autoencoders, and Other Techniques.

\subsubsection{What are the different types of techniques researched?}

\begin{table}
\centering
\begin{tabular}{L{2.5cm} L{4cm} L{5cm}}
\toprule
Method Type & Usages in Collected Literature & Percentage of All Method Usages \\
\midrule
\textbf{GANs} & \textbf{76} & \textbf{73.8\%} \\
\addlinespace
CGANs & 16 & 15.5\% \\
V/WGANs & 18 & 17.5\% \\
Other GANs & 42 & 40.8\% \\
\addlinespace
\textbf{Autoencoders} & \textbf{9} & \textbf{8.7\%} \\
\addlinespace
\textbf{Other} & \textbf{18} & \textbf{17.5\%} \\
\bottomrule
\end{tabular}
\caption[Summary of Generative Technique Usages in Collected Studies.]{Summary of generative technique usages in collected studies. Each usage is counted per study. For example, one method applied in two studies counts as two usages; two methods in one study count as two usages.}
\label{tab:rq1-model-summary}
\end{table}

\begin{figure}
\centering
\includegraphics[width=\textwidth]{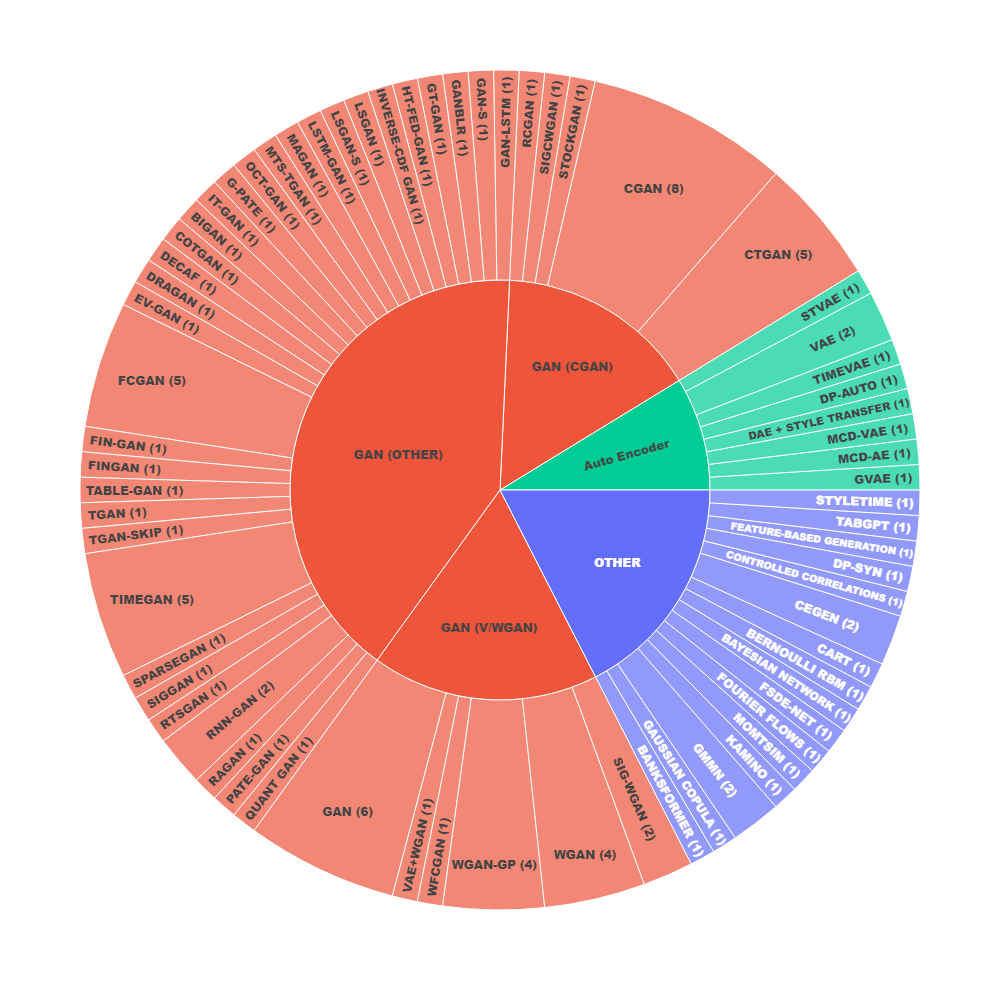}
\caption[Sunburst chart of generative models used in collected studies.]{Sunburst chart of generative models used in collected studies. The outer layer shows the specific methods (with the number of studies in parentheses); the inner layer shows the overarching architectures.}
\label{fig:models-sundial}
\end{figure}

The taxonomy in Figure~\ref{fig:models-sundial} illustrates that Generative Adversarial Networks (GANs) are by far the most heavily researched family of methods for financial synthetic data generation. Within GANs, Conditional GANs, Vanilla GANs, and Wasserstein GANs are the most prevalent variants. Autoencoders are the next most common, particularly Variational Autoencoders, while a variety of other techniques, including Generative Moment Matching Networks and transformers, have also been explored.

\textbf{Conditional GANs}: Conditional GANs (CGANs) generate data conditioned on auxiliary information, making them well suited for datasets with categorical or structured attributes. Across our collection, five main variants of Conditional GANs were assessed (Table~\ref{tab:rq1-conditional-gans}).

STOCKGAN \citep{Li2020GeneratingStreams} is designed specifically for generating realistic stock market orders by capturing historical dependencies. Compared to Variational Autoencoders and Deep Convolutional GANs, STOCKGAN showed a significant performance advantage in Kolmogorov–Smirnov distance but was not evaluated for privacy preservation or downstream ML performance.

SIGCWGAN \citep{Ni2020ConditionalGeneration} and RCGAN \citep{Hyland2017REAL-VALUEDGANS} were assessed together by \citet{Gatta2022NeuralSeries}, who found SIGCWGAN particularly promising for privacy preservation (with a perfect innovation score) but less competitive in predictive performance.

CTGAN and CTab-GAN were primarily applied to tabular credit and transaction data. CTGAN \citep{Xu2019ModelingGAN} demonstrated strong improvements over benchmarks such as TGAN in modelling mixed-type data. CTab-GAN \citep{Zhao2021CTAB-GAN:Synthesizing} further extends this by encoding mixed variables, though specific performance on financial datasets was only reported in aggregate.

\begin{table}
\centering
\begin{tabular}{l l l}
\toprule
\textbf{Model} & \textbf{Description} & \textbf{Studies} \\
\midrule
CGAN & Conditional GAN & {[}2{]} {[}6{]} {[}10{]} {[}16{]} {[}18{]} {[}27{]} {[}42{]} {[}61{]} \\
STOCKGAN & Stock Market Order CGAN & {[}19{]} \\
CTGAN & Conditional Tabular GAN & {[}29{]} {[}32{]} {[}35{]} {[}37{]} \\
SIGCWGAN & Signature Wasserstein CGAN & {[}62{]} \\
RCGAN & Recurrent Conditional GAN & {[}62{]} \\
\bottomrule
\end{tabular}
\caption{Conditional GANs used across studies.}
\label{tab:rq1-conditional-gans}
\end{table}

\textbf{Vanilla and Wasserstein GANs}: Several studies compared Vanilla GANs with Wasserstein GAN variants, often finding the latter superior for capturing realistic time series and improving predictive performance. For example, \citet{Simonetto2018GeneratingTransactions} observed that WGAN-GP outperformed other GAN variants in both statistical similarity and machine learning efficacy when generating spiking time series.

\begin{table}
\centering
\begin{tabular}{l l l}
\toprule
\textbf{Model} & \textbf{Description} & \textbf{Studies} \\
\midrule
GAN & Generative Adversarial Network & {[}5{]} {[}26{]} {[}51{]} {[}53{]} {[}56{]} {[}71{]} \\
WGAN-GP & Wasserstein GAN with Gradient Penalty & {[}8{]} {[}36{]} {[}38{]} {[}39{]} \\
SIG-WGAN & Signature Wasserstein GAN & {[}9{]} {[}51{]} \\
WFCGAN & Wasserstein Fully Convolutional GAN & {[}26{]} \\
WGAN & Wasserstein GAN & {[}26{]} {[}37{]} {[}39{]} {[}51{]} \\
VAE+WGAN & VAE Generator within WGAN & {[}39{]} \\
\bottomrule
\end{tabular}
\caption{Vanilla and Wasserstein GANs.}
\label{tab:vwgan}
\end{table}

\textbf{Other GAN Variants}: Many less common GAN variants have been explored (Table~\ref{tab:other-gans}). For example, TimeGAN \citep{Yoon2019Time-seriesNetworks} was repeatedly observed to perform well across predictive and discriminative tasks, while FCGAN showed mixed results, sometimes improving regression accuracy but often underperforming WGANs. PATE-GAN \citep{Jordon2019PATE-GAN:GUARANTEES} and G-PATE \citep{Long2021G-PATE:Discriminators} demonstrated the feasibility of achieving differential privacy with relatively limited performance degradation.

\begin{table}
\resizebox{\columnwidth}{!}{%
\begin{tabular}{@{}lll@{}}
\toprule
\textbf{Model Name} & \textbf{Description} & \textbf{Studies} \\
\midrule
TIMEGAN             & Time GAN                                     & {[}0{]} {[}21{]} {[}62{]} {[}64{]} {[}72{]} \\
FCGAN               & Fully Convolutional GAN                      & {[}5{]} {[}26{]} {[}71{]} \\
RNN-GAN             & Recurrent Neural Network GAN                 & {[}11{]} {[}57{]} \\
LSTM-GAN            & Long Short-Term Memory GAN                   & {[}13{]} \\
GANBLR              & GAN inspired by Naive Bayes and Logistic Regression & {[}15{]} \\
PATE-GAN            & Private Aggregation of Teacher Ensembles GAN & {[}17{]} \\
QUANT GAN           & Quant GAN                                    & {[}20{]} \\
DRAGAN              & Deep Regret Analytic GAN                     & {[}22{]} \\
TABLE-GAN           & Table GAN                                    & {[}30{]} \\
GAN-LSTM            & GAN-LSTM                                     & {[}34{]} \\
RAGAN               & Relativistic Average GAN                     & {[}36{]} \\
TGAN                & Tabular GAN                                  & {[}37{]} \\
TGAN-SKIP           & TGAN with Skip Connections                   & {[}38{]} \\
MTS-TGAN            & Multivariate Time Series TGAN                & {[}40{]} \\
FINGAN              & FinGAN                                       & {[}41{]} \\
OCT-GAN             & NODE-based Conditional Tabular GAN           & {[}43{]} \\
G-PATE              & Generative Private Aggregation of Teacher Ensembles & {[}44{]} \\
INVERSE-CDF GAN     & Inverse Cumulative Distribution Function GAN & {[}45{]} \\
DECAF               & Debiasing Causal Fairness GAN                & {[}46{]} \\
IT-GAN              & Invertible Tabular GAN                       & {[}47{]} \\
HT-FED-GAN          & Horizontal Tabular Federated GAN             & {[}48{]} \\
GAN-S               & Signature GAN                                & {[}51{]} \\
LSGAN               & Least Squares GAN                            & {[}51{]} \\
LSGAN-S             & Signature Least Squares GAN                  & {[}51{]} \\
EV-GAN              & Extreme-Value GAN                            & {[}54{]} \\
SPARSEGAN           & Sparse Self-Attention Guided GAN             & {[}58{]} \\
RTSGAN              & Real World Time Series GAN                   & {[}59{]} \\
MAGAN               & Multi-Attention GAN                          & {[}60{]} \\
BIGAN               & Bidirectional GAN                            & {[}61{]} \\
GT-GAN              & General Purpose Time Series GAN              & {[}63{]} \\
FIN-GAN             & FIN-GAN                                      & {[}67{]} \\
COTGAN              & Causal Optimal Transport GAN                 & {[}72{]} \\
SIGGAN              & Signature GAN                                & {[}72{]} \\
\bottomrule
\end{tabular}%
}
\caption{Other GAN architectures.}
\label{tab:other-gans}
\end{table}

\textbf{Autoencoders}: Autoencoders, especially VAEs, were primarily used for tabular or time series reconstruction. For example, \citet{Desai2021TimeVAE:Generation} introduced TimeVAE to incorporate trend and seasonality into the decoder, achieving comparable results to TimeGAN. \citet{Miok2019GeneratingDropout} evaluated Monte Carlo Dropout regularisation in VAEs and AEs, finding marginal improvements in some predictive tasks.

\begin{table}
\centering
\begin{tabular}{l l l}
\toprule
\textbf{Model} & \textbf{Description} & \textbf{Studies} \\
\midrule
VAE & Variational Autoencoder & {[}5{]} {[}71{]} \\
TimeVAE & Temporal Variational Autoencoder & {[}23{]} \\
DAE + Style Transfer & Denoising Autoencoder with Style Transfer & {[}24{]} \\
MCD-VAE & Monte Carlo Dropout VAE & {[}66{]} \\
\bottomrule
\end{tabular}
\caption{Autoencoders used across studies.}
\label{tab:aes}
\end{table}

\textbf{Other Techniques}: Some studies evaluated methods outside GANs and autoencoders, such as CEGEN, which outperformed GANs in discriminative and predictive metrics for time series \citep{Remlinger2021ConditionalGeneration}. GMMN also achieved strong results for predictive performance. Finally, TABGPT \citep{Padhi2020TabularSeries} demonstrated promising results for privacy-preserving tabular data generation.

\begin{table}
\centering
\begin{tabular}{l l l}
\toprule
\textbf{Model} & \textbf{Description} & \textbf{Studies} \\
\midrule
CEGEN & Conditional Euler Generator & {[}52{]} {[}72{]} \\
GMMN & Generative Moment Matching Network & {[}5{]} {[}62{]} \\
TABGPT & Transformer for Tabular Generation & {[}31{]} \\
Fourier Flows & Time Series via Fourier Transforms & {[}68{]} \\
Kamino & Constraint-Aware DP Synthesis & {[}74{]} \\
\bottomrule
\end{tabular}
\caption{Other generative techniques.}
\label{tab:others}
\end{table}

\subsubsection{How are different techniques used across different financial applications?}

Table~\ref{tab:methods-applications} summarises the distribution of generative techniques across financial applications. 
The synthesis of stock market data dominates the literature. Across 34 distinct generative approaches, 27 are GAN variants, including 19 outside the common Conditional, Vanilla, or Wasserstein families. TimeGAN appears most frequently, used in four studies, followed by Vanilla GAN and Conditional GAN, each applied in three studies. This shows that stock market data has been the main focus of synthetic financial data research over the last five to six years.  

A similar pattern is found in studies on currency and commodity data. Although only nine and ten studies focus on these areas, all major categories of generative models are represented. Apart from CGAN, which is applied in two currency studies, most techniques are used only once. This suggests that research in these areas is still exploratory. Some studies generate both currency and commodity datasets, reflecting their related time-series properties.  

For market orders, two studies applied Conditional GANs, including the specialised StockGAN model designed to capture order flow behaviour.  
Credit and loan data show more diversity. Three autoencoder variants were used for credit risk prediction, and two for fraud detection. However, GANs remain the most common choice overall, and they are the only type of model used for generating personal loan, credit card, marketing, churn, and economic scenario data.  

Transaction data is the only application where non-GAN techniques appear more often than GANs. Of the eleven techniques identified, six are non-GAN architectures such as transformers and statistical models, while five are GAN-based. This reflects the challenges of transactional data, such as sparsity and categorical imbalance.  
In summary, GANs dominate across almost all applications, but there is no single application and generator combination that clearly stands out. This suggests both flexibility in generative modelling and an ongoing search for the most suitable methods for different types of financial data.

\begin{table}
\small
\centering
\begin{tabular}{
    L{2cm}
    L{2cm}
    L{3cm}
    L{3cm}
}
\toprule
\textbf{Application} &
\textbf{Unique Techniques} &
\textbf{Top Categories (Count)} &
\textbf{Most Common Techniques} \\
\midrule
Stock Market &
34 &
Other GAN (19), Other (5), V/WGAN (5), CGAN (3) &
TimeGAN (4), GAN (3), CGAN (3) \\
\addlinespace
Market Orders &
2 &
CGAN (2) &
CGAN (2), STOCKGAN (1) \\
\addlinespace
Currency Exchange &
9 &
Other GAN (3), AE (2), Other (2) &
CGAN (2) \\
\addlinespace
Commodities &
10 &
Other GAN (5), V/WGAN (2) &
WGAN-GP (1), CGAN (1) \\
\addlinespace
Credit Risk &
9 &
AE (3), Other GAN (3) &
CGAN (3) \\
\addlinespace
Credit Fraud &
10 &
Other GAN (5), AE (2) &
CTGAN (2), PATE-GAN (1) \\
\addlinespace
Loan &
3 &
CGAN (2) &
CGAN (1), CTGAN (1) \\
\addlinespace
Transaction &
11 &
Other (6), V/WGAN (3) &
WGAN-GP (2), TABGPT (1) \\
\addlinespace
Marketing &
3 &
CGAN (1), V/WGAN (1) &
CTGAN (2) \\
\addlinespace
Churn &
1 &
CGAN (1) &
CTGAN (1) \\
\addlinespace
Retail Prices &
2 &
Other GAN (1), Other (1) &
TABLE-GAN (1), CEGEN (1) \\
\addlinespace
Economic Scenario &
1 &
V/WGAN (1) &
GAN (1) \\
\addlinespace
Tax &
1 &
Other (1) &
Kamino (1) \\
\bottomrule
\end{tabular}
\caption{Generative techniques by financial application.}
\label{tab:methods-applications}
\end{table}

\subsection{What evaluation methods and criteria have been used?} \label{rq:3}

This section analyses the methods used in the literature to evaluate the quality of synthetic financial data. We begin with the broader evaluation criteria applied across studies, followed by a discussion of the specific metrics used for each criterion.

\subsubsection{General Evaluation Criteria}

Table~\ref{tab:rq3-evaluation-categories} summaries the general evaluation approaches reported in the collected studies. The results in the table indicate that Statistical Similarity and Machine Learning Efficacy are the two most common evaluation criteria. In addition, 61.6\% of studies compare their methods against existing benchmarks. Privacy Preservation was explicitly assessed in only 12.3\% of the literature. Finally, one study reported only descriptive observations of the generated data rather than formal evaluation.

\begin{table}
\centering
\begin{tabular}{lcc}
\toprule
\textbf{Evaluation Criterion} & \textbf{Number of Studies} & \textbf{Percentage} \\
\midrule
Statistical Similarity    & 58 & 79.5\% \\
Machine Learning Efficacy & 48 & 65.8\% \\
Comparison to Benchmarks  & 45 & 61.6\% \\
Privacy Preservation      & 11 & 12.3\% \\
General Observation       & 1  & 1.4\%  \\
\bottomrule
\end{tabular}
\caption{Evaluation criteria used in the collected studies.}
\label{tab:rq3-evaluation-categories}
\end{table}

These findings highlight an imbalance in evaluation practices. While most research focuses on similarity to real data and downstream machine learning performance, considerably less attention is given to privacy preservation. This gap suggests opportunities for future work that more rigorously tests the privacy properties of synthetic financial data.

\subsubsection{Statistical Similarity}

A total of 58 studies in our collection evaluated synthetic financial data in terms of statistical similarity to real data. 
A common approach is to train a discriminator model to classify between real and synthetic samples, following the same principle as GAN training. Ideally, the classifier should perform no better than random guessing (i.e., an accuracy of 0.5). However, there is no consensus on the choice of classifier. For example, \citet{Yoon2019Time-seriesNetworks}, \citet{Remlinger2021ConditionalGeneration}, and \citet{Pei2021TowardsData} used LSTM-based classifiers, while others employed neural networks, logistic regression, random forests, support vector machines, or k-nearest neighbours. The lack of consistency suggests that multiple classifiers may need to be combined to strengthen evaluation results.

\begin{figure}
    \centering
    \includegraphics[width=0.6\textwidth]{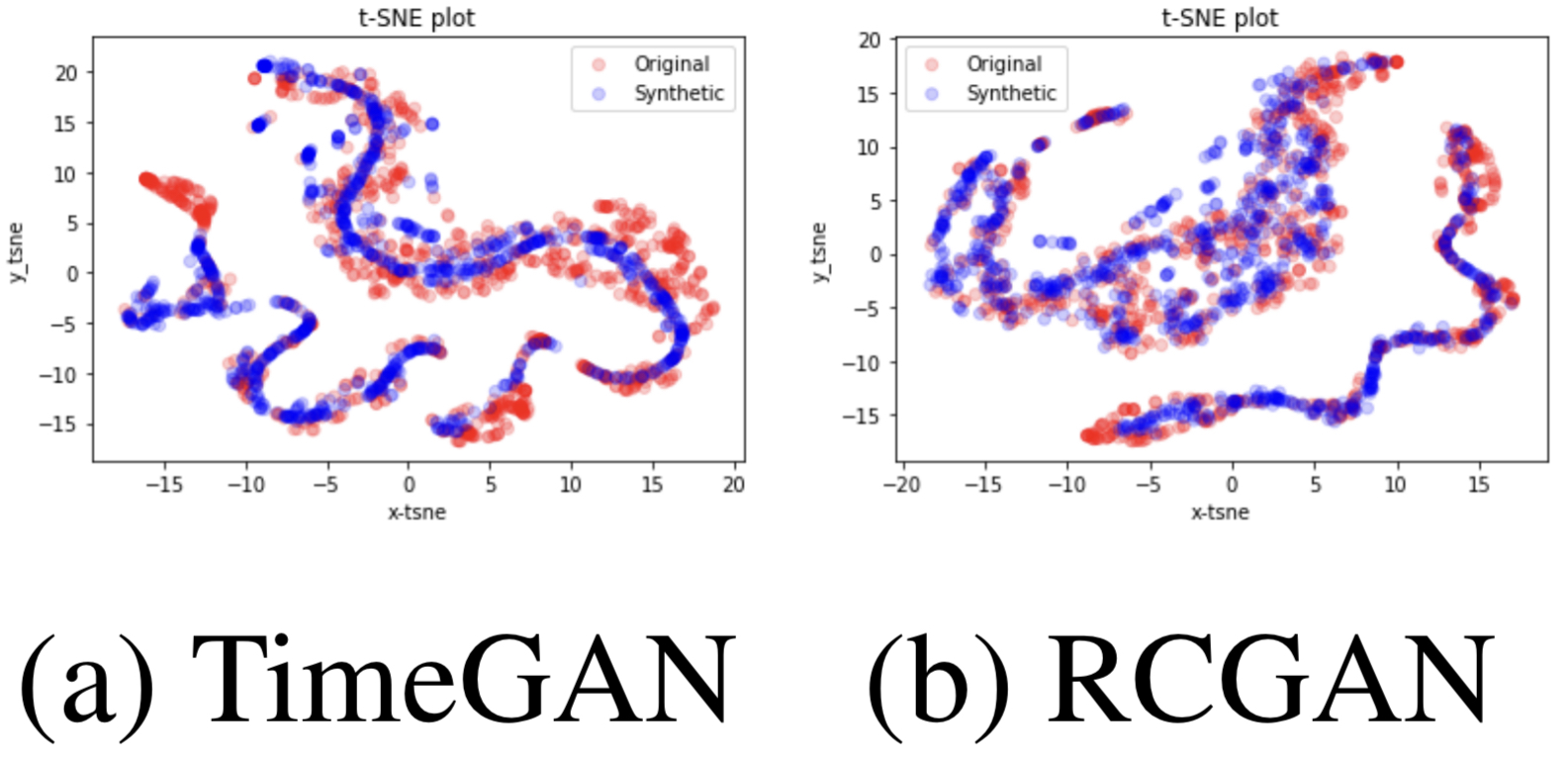}
    \caption[t-SNE plot example.]{Example of a t-SNE plot comparing the performance of two generative techniques on stock market data \citep{Yoon2019Time-seriesNetworks}.}
    \label{fig:rq3-t-sne}
\end{figure}

Visual inspection was also widely applied to assess statistical similarity. Common practices included comparing cross-correlations, distributional shapes, t-SNE plots for high-dimensional structure, and side-by-side inspection of synthetic and real time series. Visualisation improves the interpretability of results and can support the explainability of model decisions \citep{Kovalerchuk2020SurveyQuasi-explanations}. Distributional comparisons were the most frequent, with 18 studies analysing whether synthetic data preserved features such as heavy tails \citep{Dogariu2021TowardsLearning,Ljung2021SyntheticNetworks,Allouche2022EV-GAN:Networks,Karlsson2020SynthesisNetworks} or categorical frequency distributions.

t-SNE plots were used in 11 studies to visualise high-dimensional relationships, making it easier to assess whether the structural properties of real data were maintained in synthetic datasets (see Figure~\ref{fig:rq3-t-sne}). Correlation-based measures such as autocorrelation and pairwise feature correlations were also applied in 20 studies, reflecting the importance of capturing dependencies between variables in financial data.

Distance metrics provided a more formal means of comparison. The most frequently applied measures included Kullback–Leibler (KL) divergence, Jensen–Shannon (JS) divergence, Kolmogorov–Smirnov (KS) statistic, and Earth Mover (EM) distance (also known as Wasserstein distance). KL and JS divergences were often used to compare probability distributions, while EM distance was applied to both time series and tabular data. The KS statistic was the most widely used single metric, appearing in seven studies across domains including stock market, transaction, currency exchange, and credit risk data. Its broad use underscores its general applicability for assessing distributional similarity.
Overall, statistical similarity evaluations ranged from qualitative visual analysis to formal statistical tests. Table~\ref{tab:stat-methods} summarises the main approaches observed across the literature.

\begin{table}
\resizebox{\columnwidth}{!}{%
\begin{tabular}{|l|l|l|l|}
\hline
\textbf{Method} & \textbf{Variation} & \textbf{Number of Studies} & \textbf{Studies} \\ \hline
Absolute Kendall Error & Absolute Kendall Error & 1 & {[}54{]} \\ \hline
Basic Statistics &
 Basic Statistics &
 7 &
 {[}27{]} {[}33{]} {[}38{]} {[}41{]} {[}50{]} {[}52{]} {[}66{]} \\ \hline
Calmar & Calmar & 1 & {[}18{]} \\ \hline
\multirow{8}{*}{Correlation} &
 Autocorrelation Function &
 4 &
 {[}20{]} {[}26{]} {[}55{]} {[}69{]} \\
 & Correlation Ratio & 1 & {[}38{]} \\
 & Covariance & 2 & {[}16{]} {[}42{]} \\
 & Feature Correlation & 7 & {[}9{]} {[}10{]} {[}12{]} {[}35{]} {[}52{]} {[}62{]} {[}72{]} \\
 & Mirror Column Associations & 1 & {[}38{]} \\
 & Mutual Information & 1 & {[}37{]} \\
 & Pairwise Association & 1 & {[}29{]} \\
 & field correlation stability & 1 & {[}64{]} \\ \hline
Cumulative Distributions & Cumulative Distributions & 1 & {[}48{]} \\ \hline
Deep Structure Stability & Deep Structure Stability & 1 & {[}64{]} \\ \hline
Dimension Reduction Score & Dimension Reduction Score & 2 & {[}61{]} {[}62{]} \\ \hline
\multirow{6}{*}{Discriminator} & KNN & 2 & {[}62{]} {[}63{]} \\
 & LSTM & 3 & {[}21{]} {[}52{]} {[}59{]} \\
 & Logistic Regression & 1 & {[}29{]} \\
 & Not Given & 1 & {[}23{]} \\
 & Random Forest & 1 & {[}62{]} \\
 & SVM & 1 & {[}39{]} \\ \hline
DY Metric & DY Metric & 1 & {[}20{]} \\ \hline
EM Distance & EM Distance & 5 & {[}5{]} {[}20{]} {[}26{]} {[}35{]} {[}71{]} \\ \hline
\multirow{2}{*}{Feature Importance} & Global Feature Importance & 1 & {[}70{]} \\
 & Local Feature Importance & 1 & {[}70{]} \\ \hline
\multirow{2}{*}{Feature Interaction} & Global Feature Interaction & 1 & {[}70{]} \\
 & Local Feature Interaction & 1 & {[}70{]} \\ \hline
Feature-based Distance & Feature-based Distance & 1 & {[}14{]} \\ \hline
Field Distribution Stability & Field Distribution Stability & 1 & {[}64{]} \\ \hline
Frechet Inception Distance & Frechet Inception Distance & 1 & {[}31{]} \\ \hline
Hurst Index & Hurst Index & 1 & {[}55{]} \\ \hline
Jenson-Shannon & Jenson-Shannon & 3 & {[}5{]} {[}26{]} {[}35{]} \\ \hline
Joint Quantile Exceedance & Joint Quantile Exceedance & 1 & {[}53{]} \\ \hline
Kendall & Kendall & 1 & {[}33{]} \\ \hline
KL & KL & 3 & {[}5{]} {[}26{]} {[}62{]} \\ \hline
Kolmogorov-Smirnov &
 Kolmogorov-Smirnov &
 7 &
 {[}3{]} {[}5{]} {[}19{]} {[}24{]} {[}26{]} {[}56{]} {[}61{]} \\ \hline
Leverage Effect & Leverage Effect & 2 & {[}20{]} {[}67{]} \\ \hline
Local Sensitivity Analysis & Local Sensitivity Analysis & 1 & {[}70{]} \\ \hline
MDFA & MDFA & 1 & {[}69{]} \\ \hline
outlier filter & outlier filter & 1 & {[}64{]} \\ \hline
Overfitting Prevention & Overfitting Prevention & 1 & {[}64{]} \\ \hline
PCA & PCA & 1 & {[}38{]} \\ \hline
Pearson & Pearson & 5 & {[}16{]} {[}33{]} {[}37{]} {[}38{]} {[}42{]} \\ \hline
Power Spectral Density & Power Spectral Density & 1 & {[}69{]} \\ \hline
Probability Density Function & Probability Density Function & 3 & {[}2{]} {[}64{]} {[}69{]} \\ \hline
RFE & RFE & 1 & {[}8{]} \\ \hline
Similarity Filter & Similarity Filter & 1 & {[}64{]} \\ \hline
Single Value Relative Error & Single Value Relative Error & 1 & {[}8{]} \\ \hline
Spearman & Spearman & 3 & {[}16{]} {[}33{]} {[}42{]} \\ \hline
Stylized Facts & Stylized Facts & 1 & {[}67{]} \\ \hline
Uncertainty Coefficient & Uncertainty Coefficient & 1 & {[}38{]} \\ \hline
Variation Distance & Variation Distance & 2 & {[}73{]} {[}74{]} \\ \hline
\multirow{9}{*}{Visual Inspection} &
 Visual Inspection (Correlations) &
 6 &
 {[}10{]} {[}25{]} {[}27{]} {[}29{]} {[}36{]} {[}38{]} \\
 &
 Visual Inspection (Distributions) &
 18 &
 {[}2{]} {[}6{]} {[}9{]} {[}10{]} {[}19{]} {[}26{]} {[}27{]} {[}29{]} {[}30{]} {[}31{]} {[}33{]} {[}38{]} {[}41{]} {[}50{]} {[}55{]} {[}56{]} {[}63{]} {[}69{]} \\
 & Visual Inspection (PCA) & 2 & {[}24{]} {[}40{]} \\
 & Visual Inspection (Statistics) & 1 & {[}3{]} \\
 & Visual Inspection (Variance) & 1 & {[}27{]} \\
 & Visual Inspection (bitmap) & 1 & {[}39{]} \\
 & Visual Inspection (chi2) & 1 & {[}31{]} \\
 &
 Visual Inspection (t-sne) &
 11 &
 {[}0{]} {[}1{]} {[}21{]} {[}22{]} {[}23{]} {[}40{]} {[}58{]} {[}59{]} {[}63{]} {[}64{]} {[}68{]} \\
 &
 Visual Inspection (time-series) &
 13 &
 {[}0{]} {[}10{]} {[}19{]} {[}24{]} {[}27{]} {[}33{]} {[}36{]} {[}39{]} {[}40{]} {[}52{]} {[}60{]} {[}62{]} {[}69{]} \\ \hline
\end{tabular}%
}
\caption{Methods and metrics for Statistical Similarity.} \label{tab:stat-methods}
\end{table}

\subsubsection{Machine Learning Efficacy}

With the rise of machine learning in financial organisations, machine learning efficacy has become a central focus for assessing the usability of generated financial data. In this subsection, we review the metrics used to evaluate the machine learning efficacy of synthetic data generation techniques.

The predominant approach is the Train–Synthetic–Test–Real (TSTR) protocol, in which a model is trained on synthetic data and evaluated on real data. Performance is typically compared against the same architecture trained on real data. Results may be reported either as the difference between the two test scores or as both scores side by side; these presentations convey the same information. Ideally, models trained on synthetic data perform \textbf{at least} as well as those trained on real data. Variation in TSTR assessments largely stems from the choice of comparison metrics. For classification tasks, \textit{accuracy} and \textit{F1} are the most common statistics, whereas \textit{mean absolute error (MAE)} is the standout metric for regression tasks. Several studies also report \textit{precision} and \textit{recall}, the components of F1 that capture, respectively, the quality and quantity of positive predictions. For tasks such as fraud detection, recall is especially important because the aim is to identify as many fraudulent actors or transactions as possible; for credit risk assessment, precision is often preferred because the objective is to approve only loans that will be repaid.

A subset of studies evaluates synthetic financial data by training trading models on generated data and measuring their investment performance. We observe a range of portfolio and risk-adjusted metrics. \cite{Liu2022SyntheticTrading} reports annual returns alongside the Sharpe and Sortino ratios to analyse performance; these statistics indicate how an equity investment performs relative to a risk-free benchmark. Similarly, \cite{Wu2023AApproaches} uses cumulative return and the Sharpe ratio, as well as \textit{winning percentage}, defined as the fraction of trading pairs with positive returns. Average returns are also used as an efficacy metric in \cite{Zhang2019StockNetwork}.

\begin{table}
\resizebox{\columnwidth}{!}{%
\begin{tabular}{|l|l|l|l|}
\hline
\textbf{Method} &
 \textbf{Variation} &
 \textbf{Number of Studies} &
 \textbf{Studies} \\ \hline
A* & A* & 1 & {[}1{]} \\ \hline
Accuracy &
 Accuracy &
 12 &
 {[}3{]} {[}5{]} {[}11{]} {[}15{]} {[}16{]} {[}29{]} {[}32{]} {[}35{]} {[}36{]} {[}42{]} {[}71{]} {[}74{]} \\ \hline
Adjusted Rand Index & Adjusted Rand Index & 1 & {[}66{]} \\ \hline
AUC & AUC & 4 & {[}16{]} {[}22{]} {[}42{]} {[}56{]} \\ \hline
AugMAE & AugMAE & 1 & {[}1{]} \\ \hline
AUPRC & AUPRC & 1 & {[}17{]} \\ \hline
AUROC & AUROC & 3 & {[}17{]} {[}44{]} {[}46{]} \\ \hline
F1 &
 F1 &
 13 &
 {[}1{]} {[}10{]} {[}16{]} {[}32{]} {[}35{]} {[}38{]} {[}42{]} {[}43{]} {[}45{]} {[}47{]} {[}48{]} {[}68{]} {[}74{]} \\ \hline
F2 & F2 & 1 & {[}65{]} \\ \hline
\multirow{5}{*}{Investments} 
 & Average Return & 1 & {[}13{]} \\ 
 & Cumulative Returns & 2 & {[}51{]} {[}0{]} \\
 & Sharpe Ratio & 3 & {[}0{]} {[}18{]} {[}51{]} \\
 & Sortino Ratio & 1 & {[}0{]} \\ 
 
 & Winning PCT & 1 & {[}51{]} \\ \hline
Agreement Rate & Agreement Rate & 1 & {[}28{]} \\ \hline
 Volatility Clustering & Volatility Clustering & 1 & {[}26{]} \\ \hline 
MAE &
 MAE &
 14 &
 {[}1{]} {[}13{]} {[}21{]} {[}23{]} {[}37{]} {[}40{]} {[}48{]} {[}52{]} {[}57{]} {[}58{]} {[}59{]} {[}60{]} {[}63{]} {[}68{]} \\ \hline
MAPE & MAPE & 3 & {[}13{]} {[}25{]} {[}60{]} \\ \hline
MSE & MSE & 3 & {[}57{]} {[}62{]} {[}72{]} \\ \hline
MSLE & MSLE & 2 & {[}40{]} {[}54{]} \\ \hline
Observation & Observation & 1 & {[}12{]} \\ \hline
Precision & Precision & 4 & {[}10{]} {[}11{]} {[}46{]} {[}65{]} \\ \hline
R2 & R2 & 2 & {[}55{]} {[}60{]} \\ \hline
Recall & Recall & 4 & {[}10{]} {[}11{]} {[}46{]} {[}65{]} \\ \hline
Replication Errors & Replication Errors & 1 & {[}72{]} \\ \hline
RMSE & RMSE & 4 & {[}13{]} {[}18{]} {[}25{]} {[}60{]} \\ \hline
RNSE & RNSE & 1 & {[}38{]} \\ \hline
ROC & ROC & 2 & {[}35{]} {[}56{]} \\ \hline
ROC Curve & ROC Curve & 1 & {[}29{]} \\ \hline
ROCAUC & ROCAUC & 1 & {[}47{]} \\ \hline
SMAPE & SMAPE & 1 & {[}60{]} \\ \hline
\multirow{2}{*}{Visual Inspection} &
 Visual Inspection (Classifications) &
 1 &
 {[}30{]} \\
 & Visual Inspection (trading actions) & 1 & {[}0{]} \\ \hline
\end{tabular}%
}
\caption{Methods and metrics for Machine Learning Efficacy.} \label{tab:mle-methods}
\end{table}

\subsubsection{Privacy Preservation}

Although privacy preservation has received relatively little attention in the literature we collected, a variety of techniques have still been applied to assess synthetic financial datasets (Table~\ref{tab:pp-methods}). Two studies, \cite{Park2018DataNetworks} and \cite{Duan2022HT-Fed-GAN:Synthesis}, experiment with membership inference attacks to evaluate the privacy of their generative techniques. These attacks attempt to infer whether specific training data were used by testing input samples and observing whether the model makes high-confidence predictions. \cite{Duan2022HT-Fed-GAN:Synthesis} provide a detailed description of their procedure and report improved privacy when using higher values of differential privacy. \cite{Park2018DataNetworks}, in addition to using nearest neighbour evaluation methods, apply a similar approach and likewise find that higher differential privacy values reduce leakage. Both studies note that lower values can lead to data leakage, highlighting the importance of privacy testing in synthetic data evaluation and exposing a concerning gap in the literature where so few studies include privacy assessment in their experiments.

The distance between synthetic and real samples is the most widely used method for assessing privacy in synthetic financial data. Five studies detect potential violations by identifying synthetic samples that lie within a small Euclidean distance of real data points. \cite{Zhao2021CTAB-GAN:Synthesizing} also employ the Nearest Neighbour Distance Ratio (NNDR). For each synthetic sample, NNDR is calculated as the ratio between the smallest distance to a real sample and the next smallest distance. An NNDR close to 1 indicates the synthetic sample lies in a dense region of real samples, whereas a value close to 0 suggests it is very close to a single real sample and distant from others. 

Similarly, \cite{vanBree2020UnlockingAutoencoders} use the mean and standard deviation of distances between synthetic and real samples to assess how easily synthetic data could be reverse-transformed to recover the original data. They report that a high mean and low standard deviation suggest synthetic samples are generally far from the original data.

Finally, \cite{Juneja2023SyntheticAnalysis} also assess privacy preservation in their generated data, but the methods are not reported and therefore cannot contribute to a deeper understanding of evaluation techniques.

\begin{table}
\centering
\resizebox{\columnwidth}{!}{%
\begin{tabular}{llll}
\toprule
\textbf{Method} & \textbf{Variation} & \textbf{Count} & \textbf{Studies} \\
\midrule
Differential Privacy & Differential Privacy & 1 & [64] \\
Membership Attacks  & Membership Attacks  & 2 & [30] [48] \\
\multirow{3}{*}{Nearest Neighbour} 
 & Nearest Neighbour              & 5 & [29] [35] [38] [62] [73] \\
 & Distance and Standard Deviation & 1 & [25] \\
 & Nearest Neighbour Distance Ratio & 1 & [35] \\
\bottomrule
\end{tabular}%
}
\caption{Methods and metrics for Privacy Preservation.}
\label{tab:pp-methods}
\end{table}

\section{Discussion} \label{chap:discussion}

Our review collected 72 studies on the generation of synthetic financial datasets, which to our knowledge is the largest collection of its kind. From this body of work, several key research focuses emerge, alongside clear gaps requiring further study. Most notably, Generative Adversarial Networks (GANs) dominate the field, featuring in 53 of the studies, while other generative approaches appear in only one or two papers each. A similar concentration is seen in the financial applications of synthetic data generation: market data (stock, currency, and commodities), transaction data, and credit-related data (risk and fraud) are the primary areas of focus. By contrast, applications such as tax records, loans, and retail data receive little attention. Perhaps the most striking gap is the limited emphasis on privacy preservation. Only 12\% of studies evaluated privacy, compared with 66\% that assessed machine learning usability and 80\% that examined statistical similarity. This is concerning, as privacy preservation is arguably the most critical feature for synthetic data in financial institutions. A likely explanation is the absence of standardised evaluation criteria for synthetic data, particularly within financial applications.

When placed in the context of other industries, such as healthcare (discussed in Section \ref{sec:introduction}), our results are consistent. For example, \cite{Hernandez2022SyntheticReview} also find that GANs dominate and that privacy is rarely evaluated. This cross-domain trend is notable, as both healthcare and finance involve highly sensitive data. For many use cases, privacy preservation should be the primary concern when generating synthetic data. While this does not invalidate the techniques used in the studies we reviewed, it suggests that industry practitioners must take responsibility for evaluating the privacy guarantees of their own implementations, and that decision-makers should prioritise privacy assessment more explicitly.

Another notable finding is how the studies were sourced. Our database search identified just 20 relevant studies out of more than 3,000, while snowballing from these initial papers yielded an additional 52. This outcome reflects the gap between our search strategy and how financial applications of synthetic data are described in the literature. Like other systematic reviews of synthetic data generation \cite{Murtaza2023SyntheticDomain}, we combined keywords related to synthetic data generation with finance-specific terms. The generation-focused terms were broad, to capture variations in terminology, but the finance-specific terms, while not overly narrow, were required for a study to be included. What emerged from the snowball search is that many relevant studies did not explicitly describe finance as a research focus in their abstract, title, or keywords, but mentioned it only when introducing datasets used for experiments. This trend makes it difficult to systematically collect literature on synthetic financial data and highlights a limitation of relying heavily on industry-specific keywords in database searches.

Another limitation of this study was the inability to compare the performance of models across different papers. Because the collected studies investigated a wide range of use cases and employed diverse evaluation methods, overall comparisons of generative models were not possible.
To the best of our knowledge, this review provides the largest analysis and collection of research on the financial applications of synthetic data to date. As a systematic review dedicated to synthetic financial data generation, it is the first of its kind and makes a significant contribution to the literature.

Opportunities for future research highlighted by our analysis include:
\begin{itemize}
    \item expanding research into generative techniques beyond GANs, including autoencoder-based generators, Generative Moment Matching Networks, and Conditional Euler Generators,
    \item greater assessment of applications such as loan data, retail data, marketing data, and tax data,
    \item the development of a standard evaluation framework for synthetic financial data, with particular emphasis on privacy preservation.
\end{itemize}

Our focus was to build a clear understanding of how synthetic datasets can be used within the financial industry. As a result, other potential uses of generated data in finance, such as the Synthetic Minority Oversampling Technique (SMOTE) and time series forecasting, were excluded from our inclusion criteria. The rapid pace of development in generative AI makes time series forecasting an especially promising area. For example, \cite{Padhi2020TabularSeries} investigated tabular time series generation, and more recently Nixtla released TimeGPT \citep{Garza2023TimeGPT-1}, a generative pre-trained transformer designed specifically for time series forecasting with a focus on financial applications. Although the released study has clear limitations, the technique itself shows strong potential. Another emerging privacy-preserving technology with applications to finance is homomorphic encryption, which allows datasets to remain encrypted while mathematical operations are performed, with the decrypted results reflecting those operations.

\section{Conclusion} \label{sec:conclusion}

This study has presented a comprehensive review of the current state of research into the financial applications of synthetic data generation. We critically analysed the focus of research across a range of applications, generative techniques, and evaluation methods. Our findings show that market data and credit data generation have received the most attention over the past five years, while other important use cases such as tax, marketing, and retail data remain comparatively underexplored, despite their sensitivity and relevance.

Generative Adversarial Networks (GANs) dominate the field, with Conditional GANs, Vanilla GANs, and Wasserstein GANs featuring prominently, and TimeGANs widely used for market data generation. A key concern we identify is the lack of evaluation of privacy preservation within the existing literature. While attributes such as statistical similarity and machine learning usability are assessed in most studies, only a small number include methods for testing the privacy guarantees of synthetic data. We strongly encourage future research to address this gap and incorporate privacy assessment into experimental design.

We also reflect on the methodological process of this systematic review, which highlights challenges in identifying relevant research given the way financial applications of synthetic data are often reported. Building on our work, future studies could expand into alternative generative approaches, explore additional financial use cases, or examine related privacy-preserving technologies such as homomorphic encryption.

As the first systematic review dedicated to synthetic data generation for finance, this study fills a notable gap in the literature. It provides clear directions for future research and offers valuable insights for industry practitioners and decision-makers considering the adoption of synthetic data technologies.





\bibliography{sn-bibliography}

\end{document}